\documentclass[letterpaper, 10 pt, conference]{ieeeconf}  

\IEEEoverridecommandlockouts                              

\usepackage{subfigure}
\usepackage{makecell}
\usepackage{cite}
\usepackage{graphicx}
\usepackage{autobreak}
\usepackage{multirow}
\usepackage{graphics} 
\usepackage{epsfig} 
\usepackage{mathptmx} 
\usepackage{times} 
\usepackage{amsmath} 
\usepackage{amssymb}  
\usepackage{amsfonts}
\usepackage{algorithm}
\usepackage{subfigure}
\usepackage[noend]{algpseudocode}
\usepackage{algorithmicx}
\makeatletter

\newcommand{\Rmnum}[1]{\expandafter\@slowromancap\romannumeral #1@}
\makeatother
\title{\LARGE \bf
A Novel Image Descriptor with Aggregated Semantic Skeleton Representation for Long-term Visual Place Recognition
}

\author{Jiwei Nie$^{1,*}$, Joe-Mei Feng$^{2,*}$, Dingyu Xue$^{1}$, Feng Pan$^{1}$, Wei Liu$^{2}$ Jun Hu$^{2}$ Shuai Cheng$^{2}$  
\thanks{*These authors contributed equally to this work and should be considered co-first authors}
\thanks{$^{1}$Jiwei Nie, Dingyu Xue, Feng Pan is with College of Information Science and Engineering, NorthEastern Universerty, ShenYang, China}
\thanks{$^{2}$Joe-Mei Feng, Wei Liu, Jun Hu, Shuai Cheng is with Neusoft Reach Automotive Technology Company, ShenYang, China}
}

\begin{document}
\maketitle
\thispagestyle{empty}
\pagestyle{empty}
\begin{abstract} 
In a Simultaneous Localization and Mapping (SLAM) system, a loop-closure can eliminate accumulated errors, which is accomplished by Visual Place Recognition (VPR), a task that retrieves the current scene from a set of pre-stored sequential images through matching specific scene-descriptors.
In urban scenes, the appearance variation caused by seasons and illumination has brought great challenges to the robustness of scene descriptors.
Semantic segmentation images can not only deliver the shape information of objects but also their categories and spatial relations that will not be affected by the appearance variation of the scene.
Innovated by the Vector of Locally Aggregated Descriptor (VLAD), in this paper, we propose a novel image descriptor with aggregated semantic skeleton representation (SSR), dubbed SSR-VLAD, for the VPR under drastic appearance-variation of environments. 
The SSR-VLAD of one image aggregates the semantic skeleton features of each category and encodes the spatial-temporal distribution information of the image semantic information. 
We conduct a series of experiments on three public datasets of challenging urban scenes.
Compared with four state-of-the-art VPR methods- CoHOG, NetVLAD, LOST-X, and Region-VLAD, VPR by matching SSR-VLAD outperforms those methods and maintains competitive real-time performance at the same time.
\end{abstract}

\section{Introduction}
Simultaneous Localization and Mapping (SLAM) can compute relevant poses while a autonomous vehicle moving in a new place.
When re-visited this place, the pre-stored poses could be utilized to adjust all the other poses of the vehicle in this environment.
One of the most critical tasks in this procedure is to locate current place from the visited places correctly. 
Visual Place Recognition (VPR) is a component of SLAM to accomplish this task.

VPR retrieves the image which is (almost-)identical to the current scene from the pre-stored images in the reference set according to visual cues\cite{Region-VLAD}.
Generally speaking, the feature-based VPR methods store a image with a its corresponding descriptor.
When performing retrieval, the similarity between images can be calculated directly by matching their descriptors.
Some hand-crafted local descriptors, such as SIFT \cite{SIFT}, SURF \cite{SURF}, BRIEF \cite{calonder2010brief}, BRISK \cite{leutenegger2011brisk}, have shown excellent performance in VPR. 
However, in the scenes where the appearance varies drastically, such as under different day time and seasons, these descriptors invalidate easily.
In the contrast, semantic information of an image is appearance-invariant.
The semantic graph model could be considered as an descriptor form to represent scenes with semantic segmentation results.
The graph-based method in \cite{x-view,guo2021semantic} encodes the spatial relationship between each semantic object in one image, and combines the odometry information to construct a 3D graph model to ensure the temporal consistency of the descriptors.
However, this method ignores the distribution of different semantic objects with the same category and the correlation between the overall semantic regions with different categories.
Furthermore, it needs to employ additional odometry information, which is unavailable in several situations, to ensure that the descriptors can encode the temporal information.
Lost-X\cite{garg2018lost} utilizes the feature map generated by CNN and the semantic segmentation map to encode the spatial layout of the semantic elements. However, it still ignores the temporal relationship among the adjacent frames. And it needs a second stage to refine the matching results.
Other semantic-based methods \cite{yu2018vlase, benbihi2020image} utilize the semantic segmentation boundary to encode the image, while this is seriously affected by the quality of the segmentation. 
%

In this paper, a novel appearance-invariant semantic image descriptor, named SSR-VLAD, is proposed.
SSR-VLAD consists of two factors- the spatial distribution between the semantic objects and the temporal information of adjacent sequential frames.
Unlike prior works as \cite{x-view, guo2021semantic}, in our method, the spatial distribution refers to the distribution of pixels with same semantic category in the image, and the relationship between different categories. 
SSR-VLAD encodes the spatial distribution through extracting the semantic skeleton of each category into a local descriptor of one image.
Then, referring to the idea of Vector of Locally Aggregated Descriptor (VLAD) \cite{VLAD}, the local semantic skeleton representation (SSR) of each category are aggregated into a fixed-dimensional global descriptor to represent the whole image. 
Moreover, the temporal part can be considered into SSR-VLAD- if given the database consisting of time-and-spatial sequential images, then considering simultaneously the adjacent frames can effectively avoid a false-true-singleton-outlier frame with high similarity score, because the database contains time-and-spatial sequential images, the similarities between the query image and the around the ground truth frame in database should also be high.
SSR-VLAD encodes the temporal information of each frame in database by taking average of the spatial part of SSR-VLAD of three images- the image, its previous- and later-images as the descriptor of each frame. 
While matching, since the spatial and the temporal information is included, a frame-to-frame matching alone is needed to measure the similarity between two images instead of frame-to-sequence matching, which is usually used in sequence-based methods \cite{garg2021seqnet}.

The main contributions of this article are briefly summarized as follows.
\begin{itemize}
%
    \item We propose a novel semantic image descriptor, called SSR-VLAD, for long-term visual place recognition. 
    This descriptor refers to the local feature aggregation idea of VLAD, and aggregates robust local semantic skeleton features into a fixed-dimensional global descriptor, so that the time consumption of image matching does not change with the complexity of the image.
    
	\item We present a spatial-temporal feature aggregation method. The descriptor can encode the semantic object distribution within and between categories. 
	Furthermore, for the reference image, the sequential information of the adjacent frames can also be encoded into the descriptor of this image. 
	With this way, the robustness and accuracy of the descriptor can be enhanced in long-term scenarios.
	
	\item We evaluate our proposed method with three SOTA VPR methods- CoHOG,NetVLAD, and Region-VLAD, in three publicly available datasets, SYNTHIA, RobotCar and Extended-CMU Season datasets which are often used to evaluate the performance of the VPR technology in the scenes with dramatic appearance changes. 
	The result shows that the describing ability of SSR-VLAD achieves state-of-the-art performance towards long-term urban scenarios with promising real-time performance.
\end{itemize}
The rest of the paper is organized as follows. In Section \ref{section2}, we introduce the related works about the global/local descriptors for VPR and semantic-based VPR approaches. Section \ref{section3} illustrates the extraction of the SSR-VLAD in detail. Experiment are detailed in Section \ref{section4} and results are discussed in Section \ref{section5}. Finally, this article is concluded in Section \ref{section6}

\section{Related Work}
\label{section2}
\subsection{Visual Place Recognition}
Most of the modern VPR methods performs feature matching based on the scene or image characteristics to calculate the similarity to search the visited scene in a database. 
In several early VPR methods, researchers focused on some appearance-based methods
\cite{bampis2018fast,FAB-MAP,DBoW2,tsintotas2018seqslam,tsintotas2021modest}, such as FAB-MAP \cite{FAB-MAP}, ORB-SLAM \cite{mur2015orb} with DBoW \cite{DBoW2}. 
These methods present good performance for indoor or short-term outdoor scenes, which are appearance-invariant.
However, as VPR is applied to a wider range of long-term scenarios, dramatic changes in appearance often result in invalidation of  appearance-based methods.
In order to solve the long-term problem, an improved appearance-based methods, named CoHOG \cite{zaffar2020cohog}, was proposed. It is a computation-efficient Histogram-of-Oriented-Gradients (HOG) based image descriptor. 
The extraction of image-entropy-based regions-of-interest (ROI) and regional-convolutional matching method maintain performance of CoHOG in changing environments. 

End-to-end VPR approaches such as NetVLAD \cite{netvlad} is a convolutional neural network (CNN), which introduces the idea of VLAD into the VPR network for the first time and trains a deep neural network to extract robust features of images and aggregate the features extracted by the network. 

Although NetVLAD was proposed in 2016, it remains state-of-the-art in several VPR benchmarks. 
Since the end-to-end network is scene-specific and requires a lot of training data, some researchers propose to extract some advanced features that are resistant to environmental changes from some general pre-training networks.
This information is mainly represented in three forms, which are the feature map of the middle convolutional layer, the semantic segmentation image, and the bounding box of detected object.
For example, Region-VLAD \cite{Region-VLAD} extracts the regional features from the middle layer of pre-trained CNN. And it is computation-efficient and environment-invariant.
In this article, we focus on utilizing semantic segmentation images for VPR. 
The semantic segmentation images represent the categories and spatial layout of the objects in the scene, and as mentioned above semantic segmentation images are relatively stable among changing scenes.
\cite{arandjelovic2014visual, mousavian2015semantically}. 

\subsection{Image Descriptors}
No matter which method mentioned above is used, there will be a descriptor designed to describe the characteristics of the image.
For the traditional appearance-based VPR methods, some hand-crafted local descriptors
\cite{SIFT,SURF,calonder2010brief,leutenegger2011brisk} are proposed to describe the key point features in the image, such as SIFT\cite{SIFT} and SURF\cite{SURF}.
Although these descriptors are rotation-, illumination-, and scale-invariant, they are not robust to changes in appearance.

Moreover, when retrieving images, calculating the descriptor correspondence keypoint-wise seriously affect the efficiency and accuracy of matching. Thus tree-structured searching algorithm based on proposed key-point-descriptor such as DBoW \cite{DBoW2} are proposed.
While building the searching tree, DBoW clusters all the given descriptors by k-means and repeats k-means several times to form a tree. For a query image, a vector will be formed by grouping all its descriptors according to the pre-trained DBoW.
Another method to speed up the image retrieval is to aggregate the local descriptors of images to form a fixed-dimensional global descriptor such as VLAD\cite{VLAD}. Through VLAD, all descriptors of an image are aggregated into one global descriptor by summing up the descriptor-residues, and thus the similarity of two images can be calculated simply by comparing the global descriptors of the two images.
DBoW considers purely the coordinate-wise distribution of the descriptors, VLAD on the other side takes spatial-relations of the descriptors into consideration. In our case, since the skeletons of the semantic segmentation represent the spatial-relations of objects in an image, using VLAD to aggregate the descriptors can retain more information of the image.

In recent years, several semantic descriptors have been proposed to solve the problem of appearance-changes.
In \cite{arandjelovic2014visual,mousavian2015semantically}, descriptors that integrate semantic information and scene appearance information are designed.
However, while facing scenes with dramatic changes in appearance, such methods may still fail.  Furthermore, they ignore the shape and spatial layout of semantic objects. 
In order to describe the spatial layout of semantic elements, \cite{x-view} and \cite{guo2021semantic} extract the center of each segmented region as a node in a graph to construct a descriptor. 
However, they ignore the relationship between elements of the same category and the relationship among different categories.

In addition, it can effectively avoid the occurrence of individual outliers to consider the semantic consistency of adjacent frames while retrieving image from a database consisting of time- and spatial- sequential images. 
In \cite{x-view,guo2021semantic}, they introduce the odometry information to connect the single semantic graph to implement this work. 
These methods require a complex frame-sequence matching and the odometry, which is not always available.
There are also several methods make use of edges of semantic segmented regions to describe the image\cite{yu2018vlase,benbihi2020image}. However, these methods rely on  good and stable semantic segmentation results.
The most similar work to this article is LOST-X. Its first stage combines the feature maps extracted by CNN and the results of semantic segmentation to calculate the descriptors of each category and finally aggregate them into a global descriptor. In the second stage, the matching results of the first stage are refined by weighted matching of key points based on semantic consistency. However, it also ignores the information of adjacent frames.
Different from LOST-X, we propose a novel semantic image descriptor. 
Firstly, we extract semantic skeleton to encode the descriptor of each category, and refer to the idea of VLAD to aggregate the descriptors of each category. 
Secondly, we present to encode the temporal relationship into the descriptor of each reference frame. 
It is worth to note that, this way only need once frame-frame matching instead of a complex frame-sequence matching.
Finally, a fixed dimension global image descriptor is obtained which encoded the information of the shape, spatial layout information of semantic elements and sequential temporal relationship.

\begin{figure}
	\includegraphics[scale=0.18]{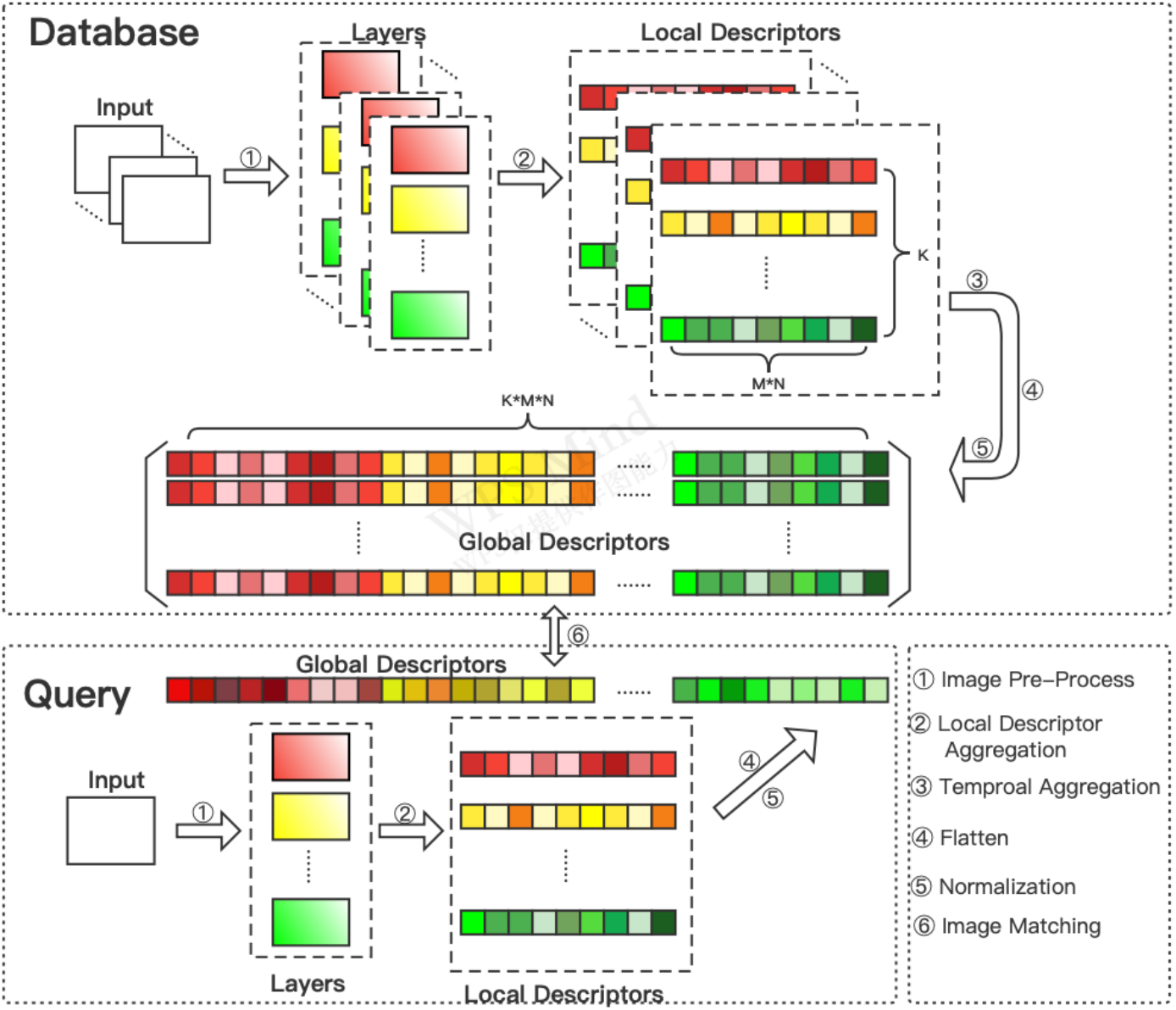}
	\caption{System Overview. The upper block and lower left block illustrates the extraction process of the image descriptor in the database and the query image respectively. The lower right block illustrates the meaning of each number symbol in the figure. The specific implement of each step is detailed in Section \ref{section3}}
	\label{fig1}
\end{figure}

\section{SSR-VLAD} 
\label{section3}
\subsection{Image Preprocessing}
As we all know, no semantic segmentation algorithm can achieve $100\%$ segmentation accuracy. In order to minimize the impact of mis-segmentation, a series of refinement processing on the semantic segmentation results are carried out. 
Firstly, the segmentation map is processed layer-by-layer according to the category, that is, the pixels with the same category label are divided into a same layer, which is a binary image.
It is worth noting that when layering is performed, the pixels labeled as  dynamic objects need to be ignored. 
In addition, the categories that are often confused are merged, such as in our case in Section \ref{section4} `wall' and `fence'.
The specific category being used is illustrated in Section \ref{section4}.
Then, for each layer, image morphology processing such as dilation and erosion are performed to eliminate noise and connect adjacent regions.
Finally, in order to further reduce the influence of mis-segmentation on the subsequent extraction of the skeleton, some small ‘holes’ contained in the enclosed segmented region are filled, and some small segmented areas that are still independent of the whole after the previous processing are removed.
The last two processing steps are very important to improve the robustness of SSR-VLAD, which is proved in the experiments.
\subsection{Local Semantic Descriptor}
After getting $K$ categorised layers of an image, denoted as $\mathbf{L} = \{l_1, l_2, \cdots, l_{K}\}$, we need to extract a set of local descriptors for each layer.
To this end, for one layer $l_i$, a skeleton is extracted. 
The endpoints and intersection-points of the skeleton can be obtained and are regarded as the key-points $\mathbf{P} = \{p_1, p_2, \cdots, p_{|key-points|}\}$ of this layer. $|key-points|$ is the total number of key-points. 
Still another, the center point $p_c$ of $l_i$ can be calculated according to Eq (\ref{eq1}).
\begin{equation}
\label{eq1}
P_{c} = (X_{c},Y_{c})=(\frac{\sum_{X}X\mathcal{I}_{l_i}(X,Y)}{\sum_{(X,Y)}\mathcal{I}_{l_i}(X,Y)}, \frac{\sum_{Y}Y\mathcal{I}_{l_i}(X,Y)}{\sum_{(X,Y)}\mathcal{I}_{l_i}(X,Y)}), 
\end{equation}
where $X$ and $Y$ represent the horizontal and vertical coordinates of the pixels $(X,Y)$ on the image and subscript $c$ means the center point, and $\mathcal{I}_{l_i}(X,Y)=1$ if pixel $(X,Y)\in {l_i}$. 
For the center point and all key-points of each layer, Shape context \cite{shape-context} is used to calculate a fixed-dimensional histogram descriptor.
Shape context takes one point $p_i$ as a reference point and $N$ concentric circles are established at logarithmic distance intervals in a local area where $p_i$ is the center of these circles and $R$ is the radius. This area is equally divided along the circumferential direction $M$ to form a target-shaped template. 
The statistical distribution histogram of the point distribution in each sector is the descriptor of the $p_i$ and the dimension of this descriptor is $1\times(M\times{N})$.
\subsection{Spatial-temporal Aggregating}
Refer to the aggregation idea of VLAD, the descriptors of all key points $\mathbf{d} = \{d_1, d_2, \cdots, d_{|key-points|}\}\in\mathrm{R}^{M\times{N}}$ are aggregated into a layer descriptor. Different from VLAD, we use the center of the semantic segmentation area of this layer instead of the center of K-means clustering.
The procedure of aggregation can be formulated as Eq (\ref{eq2}),
\begin{equation}
\label{eq2}
\mathbf{V}(k)=\sum_{i=1}^{|key-points|_k} \left(d_{i}-d_c\right)/\|\sum_{i=1}^{|key-points|_k} \left(d_{i}-d_c\right)\|_2,
\end{equation}
$k$ represents the $k_{th}$ layer and $|key-points|_k$ represents that there are $|key-points|_k$ key-points in the $k_{th}$ layer, the final descriptor $\mathbf{V}(k)$ of the $k_{th}$ layer can be obtained.
The obtained layer-descriptor $\mathbf{V}(k)$ encodes the 2D spatial layout relationship of the semantic objects of the $k_{th}$ category.

In order to encode the relationship among different category, we concatenate all of the layer descriptors and normalize it, and thus 
after the above steps, for a query image, its global descriptor is calculated into a fixed-length $\mathbf{L} = K\times M\times N$ descriptor vector $D$.

Another issue is to adding temporal information into the description. Given a query image $I_{query}$, a pre-stored sequential database $\mathbf{I} = \{I^1_{refer}, I^2_{refer}, \cdots, I^n_{refer}\}$ as well as their corresponding descriptor $D_{query}$ and $\mathbf{D} = \{D^1_{refer}, D^2_{refer}, \cdots, D^n_{refer}\}$.
While retrieving among a sequential database, in addition to the truly matching frame, there may be several individual frames that are very similar to the query image. We call these individual frames outliers.

In order to effectively reject these outliers, the normal strategy is to perform frame-sequence matching, or to average the matching results of adjacent frames.
In this article, we propose to encode the information of adjacent frames into the descriptor of one reference frame as follows, 
\begin{equation}
\label{similarity}
D^i_{refer}=\\
\sum_{f=-t}^{t} D^{i+f}_{refer},
\end{equation}
where t is the scope of the adjacent frames, so that only one round of frame-wise matching is required without additional calculations.
\subsection{Image Matching}
While performing retrieving, the similarity score between the $i_{th}$ reference image $I^i_{refer}$ and the current query image is represented as the inner product of their descriptor: 
\begin{equation}
\label{similarity}
Score<I^i_{refer}, I_{query}>=\\
 \frac{\mathbf{D}^i_{refer} \cdot \mathbf{D}_{query}}{\left\|\mathbf{D}^i_{refer}\right\| \times\left\|\mathbf{D}_{query}\right\|} 
\end{equation}
The score expresses the degree of similarity between two images, and the reference frame whose score exceeds a certain threshold can be regarded as a candidate matching frame.
For the query image $I_{query}$, the matched candidate frames could be obtained by ranking the images in database based on the similarity score. 
For a fixed segmentation category, the time complexity of matching an image pair is fixed. It will not change with the scene.

\section{Experimental Setup}
\label{section4}
\subsection{Datasets}
In order to evaluate the performance of our proposed algorithm, we utilize three publicly available long-term urban datasets: SYNTHIA\cite{ros2016synthia}, Oxford Robot Car\cite{RobotCar} and Extended-CMU Season datasets\cite{CMU,Sattler2018CVPR}.
\subsubsection{SYNTHIA}
SYNTHIA\cite{ros2016synthia} provides a synthetic dataset of urban scenes with pixel-level semantic annotations. 
This virtual city dataset includes most of the city elements, environmental changes in the four seasons, and simulates different lighting changes. 
Therefore, it is particularly suitable for verifying the scene description ability of our proposed descriptor in urban scenes with drastic changes in appearance.
\subsubsection{Oxford Robot Car}
Oxford Robot Car\cite{RobotCar} is also referred to as RobotCar dataset, which records the urban scene data of Oxford, England close to 1000Km in one year. It contains the data of the cross-seasons and day-night in same scenes.
Furthermore, some images in the RobotCar Seasons dataset have large motion blur and low image quality
\cite{Sattler2018CVPR}.
Because it provides GPS and INS ground truth, in addition to VPR, it is often used to evaluate the accuracy of long-term visual localization.
\subsubsection{Extended-CMU Season}
Extended-CMU Season is an extension of the CMU Season dataset, which adds the pose information of all conditions. This means that it can be used to verify the VPR algorithm. 
It records the scenes of city, suburban and park in different seasons of the year and under different lighting conditions.

In this article, we extract a sub-sequence from each of these datasets to evaluate SSR-VLAD.
For the SYNTHIA and RobotCar, we follow the sub-datasets abstracted in \cite{benchmark} which are regarded as benchmarks for VPR evaluation. And the ground-truth can be available\footnote{https://github.com/MubarizZaffar/VPR-Bench}. The number of query and reference images are 812, 910 and 191, 191 separately. 
For Extended-CMU Season, we select a sub-sequence of the urban scene, called Slice7, which the query sequence is recorded in 2010.10.01 and the database is recorded in 2011.11.21. This sub-sequence consists of 175 query and 190 reference images
The calculation of the ground-truth follows the criteria mentioned in \cite{benchmark}. 
According to the pose provided by the dataset, we manually select $\pm5m$ as the threshold. The reference images with a distance of less than 5m from the query image are considered to be ground-truth.
\subsection{Semantic Segmentation}
In order to extract the semantic segmentation results, we utilize an open-source CNN presented in\cite{Larsson_2019_CVPR}\footnote{https://github.com/maunzzz/cross-season-segmentation}, which is pre-trained on the Cityscapes dataset\cite{cordts2016cityscapes} and fine-tuned on RobotCar and CMU datasets.
As mentioned above, SYNTHIA dataset has provided the pixel-level annotations of each image. 
Therefore, we can directly obtain the semantic segmentation results of SYNTHIA dataset, instead of using CNN to extract.
\subsection{Baseline compared}
In order to verify the advancement and effectiveness of the proposed method, we take several previous SoTA VPR technologies(i.e. CoHOG, NetVLAD, Region-VLAD and LOST-X) as compared methods. 
CoHOG is a state-of-the-art appearance-based VPR method. 
NetVLAD and Region-VLAD are two CNN-based variants of VLAD, which have state-of-the-art performance on the evaluation datasets used in this paper.
LOST-X is a state-of-the-art semantic-based VPR method and very similar to SSR-VLAD.
We use them to measure the representation ability of SSR-VLAD.
In addition, CoHOG and Region-VLAD are two lightweight models that can run with low computing power and ensure good recognition results at the same time.
In this paper, we utilize these two lightweight methods to compare and evaluate the computational cost of SSR-VLAD. 
\subsection{Evaluation metrics}
There are lots of metrics introduced in \cite{lowry2015visual,benchmark} to evaluate the performance of VPR technologies.
In order to analyze the recognition ability, in this paper, we use the AUC-PR curve to compare the performance among the compared VPR technologies graphically. 
In addition, Recall@100\%Precision, RecallRate@N ($N$ = 1), AUC value are utilized to analyze the evaluation results quantitatively. 
Furthermore, we test SSR-VLAD on embedded platform and PC platform respectively and count the computational cost of SSR-VLAD in feature extraction, descriptor calculation and similarity calculation modules. 

\subsection{Setup, Implementation Details}\label{1111}
In this paper, we prepare two platforms for experiments. 
The one is a desktop with an Intel Core i7-10700KF (16 cores @ 2.70GHz), 16Gb RAM, and a NVIDIA RTX3070 GPU. 
The other one is an embedded platform with one TI Jacinto™7 TDA4VM and 2Gb RAM. 
In order to run SSR-VLAD in the embedded platform, we code it in the C++ programming language.
The open-source codes of all compared methods are implemented with Python.
It should be noted that the running time with Python-codes and  C++-codes can be different, and 
therefore, it should be emphasized that we do not intend to prove that SSR-VLAD is a quicker than the other methods. We just show the performance of SSR-VLAD in terms of running time by comparing with the existing SOTA lightweight models.
The parameters of all compared methods follow the default settings of \cite{benchmark}. 
For SSR-VLAD, we set $N = 5$, $M = 12$, and $t = 3$ which are same symbols mentioned in Section \ref{section3}.
For the semantic categories used, we have adjusted the default categories on Cityscapes\cite{cordts2016cityscapes} and SYNTHIA\cite{ros2016synthia} more specifically, see Table \ref{table2} for details.
\begin{table}[]
    \caption{Summary list of the used semantic categories}
    \label{table2}
    \begin{center}
        \begin{tabular}{ccc}
        \hline
        Index & Cityscapes\cite{cordts2016cityscapes} & SYNTHIA\cite{ros2016synthia} \\ \hline
        0     &{'Ground', 'Road'}  &'Road'    \\
        1     &'sidewalk'     & 'Sidewalk'   \\
        2     &{'Building', 'wall', 'Fence'}   &{'Building', 'Fence'}    \\
        3     &\makecell[c]{'Pole', 'Pole Group', 'Traffic Sign',\\ 'Traffic light'}  &{'Pole', 'Traffic Sign'}      \\
        4     & {'Vegetation', 'Terrain'}&    {'Vegetation'}     \\
        5     &     'Sky'       &  'Sky'       \\
        6     &    'Guard Rail'        &    -     \\
        7     &      'Parking'      &   -  \\ \hline
        \end{tabular}
    \end{center}
\end{table}

We conduct three experiments in order to evaluate the performance of SSR-VLAD. 
In first experiment, All methods to be compared are run on the same laptop platform with the three datasets one after another. 
Then, we qualitatively and quantitatively analyze the performance by comparing the evaluation metrics of different compared methods on these datasets.
In second experiment, we compare SSR-VLAD with two lightweight model, CoHOG and Region-VLAD in Robotcar dataset, which the size of each image is 1024*1024. 
And the computational cost of three main modules of VPR technology: feature extraction, encoding and matching are compared for evaluation. 
We run SSR-VLAD on the laptop and embedded platform. 
But for CoHOG and Region-VLAD, we only run them on the laptop because our embedded platform cannot build the Python project. 
Therefore, we only compare the computational cost of different methods in laptop. 
And the results of SSR-VLAD on the embedded platform are utilized to prove that the computational consumption of our algorithm can meet the computational requirements of common visual SLAM systems, such as ORB-SLAM2, on embedded devices.
SSR-VLAD encodes the spatial layout information of scene elements through semantic skeleton map. However, due to generalization and occlusion, the boundary of the semantic segmentation area is very unstable, which directly affects the shape of the skeleton image and the position of the skeleton point.
Therefore, in the final experiment, we aim to evaluate the robustness of SSR-VLAD under different degrees of noise. By add different degrees of noise to the extracted skeleton on the pixel coordinates to simulate the changes of the skeleton on the pixel coordinate caused by the unstable semantic segmentation boundary.
Specifically, we randomly adjust the horizontal and vertical coordinates of the extracted skeleton nodes in different interval $\delta = {\pm 25, \pm 50, \pm 100, \pm 150}$.
By matching a set of original images with their noisy images, and analyzing the results of the metric Recall@1, we can evaluate the robustness of the semantic skeleton when the boundary of the semantic segmentation is unstable.

\begin{table*}[]
\caption{The experiment results of all compared methods in three datasets}
\label{table1}
\begin{center}
    \begin{tabular}{|cccccccc|}
\hline
\multicolumn{8}{|c|}{Intel Core i7-10700KF (16 cores @ 3.80GHz), 16Gb RAM, a NVIDIA RTX3070 GPU} \\ \hline
\multicolumn{1}{|c|}{\multirow{2}{*}{Datasets}} & \multicolumn{1}{c|}{\multirow{2}{*}{Metrics}} & \multicolumn{6}{c|}{VPR Technology} \\ \cline{3-8} 
\multicolumn{1}{|c|}{}&\multicolumn{1}{c|}{}&\multicolumn{1}{c|}{CoHOG\cite{zaffar2020cohog}}&\multicolumn{1}{c|}{NetVLAD\cite{netvlad}}&\multicolumn{1}{c|}{Region-VLAD\cite{Region-VLAD}}&\multicolumn{1}{c|}{Lost-X\cite{garg2018lost}}&\multicolumn{1}{c|}{SSR-VLAD} &\makecell[c]{SSR-VLAD\\(No pre-process)}\\ \hline
\multicolumn{1}{|c|}{\multirow{3}{*}{CMU\_Season\cite{CMU}}} & 
\multicolumn{1}{c|}{Recall@100\%Precision} & \multicolumn{1}{c|}{3.1} & \multicolumn{1}{c|}{8.1} & \multicolumn{1}{c|}{10.6}& \multicolumn{1}{c|}{2.6}& \multicolumn{1}{c|}{\textbf{51.8}} &6.9 \\
\multicolumn{1}{|c|}{} & \multicolumn{1}{c|}{RecallRate@1} & \multicolumn{1}{c|}{92.4} & \multicolumn{1}{c|}{90.9} & \multicolumn{1}{c|}{91.4} & \multicolumn{1}{c|}{\textbf{92.7}}& \multicolumn{1}{c|}{81.7} & 49.7 \\
\multicolumn{1}{|c|}{} &\multicolumn{1}{c|}{AUC} &\multicolumn{1}{c|}{0.896} &\multicolumn{1}{c|}{0.931} &\multicolumn{1}{c|}{0.904} &\multicolumn{1}{c|}{0.918}&
\multicolumn{1}{c|}{\textbf{0.961}} & 0.696 \\ \hline
\multicolumn{1}{|c|}{\multirow{3}{*}{RobotCar\cite{RobotCar}}} &
\multicolumn{1}{c|}{Recall@100\%Precision} &\multicolumn{1}{c|}{9.3} & \multicolumn{1}{c|}{\textbf{80.2}} &\multicolumn{1}{c|}{7.2} &\multicolumn{1}{c|}{9.6} & \multicolumn{1}{c|}{38.5} & 14.8 \\ 
\multicolumn{1}{|c|}{} & \multicolumn{1}{c|}{RecallRate@1} & \multicolumn{1}{c|}{50.8} & \multicolumn{1}{c|}{\textbf{97.9}} & \multicolumn{1}{c|}{87.4} & \multicolumn{1}{c|}{92.1}& \multicolumn{1}{c|}{57.7} & 32.8 \\
\multicolumn{1}{|c|}{} & \multicolumn{1}{c|}{AUC} & \multicolumn{1}{c|}{0.620} & \multicolumn{1}{c|}{\textbf{0.997}} & \multicolumn{1}{c|}{0.944} & \multicolumn{1}{c|}{0.824}& \multicolumn{1}{c|}{0.855} & 0.604 \\ \hline
\multicolumn{1}{|c|}{\multirow{3}{*}{SYNTHIA\cite{ros2016synthia}}} &
\multicolumn{1}{c|}{Recall@100\%Precision} &\multicolumn{1}{c|}{2.3} & \multicolumn{1}{c|}{61.6} &\multicolumn{1}{c|}{2.0}  &\multicolumn{1}{c|}{-}& \multicolumn{1}{c|}{96.7} & \textbf{98.9} \\ 
\multicolumn{1}{|c|}{} & \multicolumn{1}{c|}{RecallRate@1} & \multicolumn{1}{c|}{69.7} & \multicolumn{1}{c|}{\textbf{99.3}} & \multicolumn{1}{c|}{62.0} & \multicolumn{1}{c|}{-}& \multicolumn{1}{c|}{99.6} & \textbf{99.9} \\
\multicolumn{1}{|c|}{} & \multicolumn{1}{c|}{AUC} & \multicolumn{1}{c|}{0.741} & \multicolumn{1}{c|}{0.998} & \multicolumn{1}{c|}{0.605}& \multicolumn{1}{c|}{-}& \multicolumn{1}{c|}{\textbf{0.999}} & \textbf{0.999} \\ \hline

\end{tabular}
\end{center}
\end{table*}

\begin{table*}[]
\caption{Results of computational cost experiments}
\label{table3}
\begin{center}
    \begin{tabular}{|cc|ccc|c|}
        \hline
    \multicolumn{2}{|l|}{\multirow{2}{*}{}} &
      \multicolumn{3}{c|}{\begin{tabular}[c]{@{}c@{}}Intel Core i7-10700KF (16 cores @ 3.80GHz), \\ 16Gb RAM,  a NVIDIA RTX3070 GPU\end{tabular}} &
      \begin{tabular}[c]{@{}c@{}}One TI Jacinto™7 TDA4VM \\ and 2Gb RAM\end{tabular} \\ \cline{3-6} 
    \multicolumn{2}{|l|}{} &
      \multicolumn{1}{c|}{CoHOG\cite{zaffar2020cohog}} &
      \multicolumn{1}{c|}{Region-VLAD\cite{Region-VLAD}} &
      SSR-VLAD &
      SSR-VLAD \\ \hline
    \multicolumn{1}{|c|}{\multirow{2}{*}{\begin{tabular}[c]{@{}c@{}}Feature \\ Encoding\end{tabular}}} &
      \begin{tabular}[c]{@{}c@{}}Feature Extraction\\  (ms)\\ (Max/Average/Min)\end{tabular} &
      \multicolumn{1}{c|}{\multirow{2}{*}{240.89/105.73/92.29}} &
      \multicolumn{1}{c|}{1478.75/1250.32/1134.92} &
      159.61/105.57/186.29 &
      709/672.658/639 \\ \cline{2-2} \cline{4-6} 
    \multicolumn{1}{|c|}{} &
      \begin{tabular}[c]{@{}c@{}}Descriptor Encoding \\ (ms)\\ (Max/Average/Min)\end{tabular} &
      \multicolumn{1}{c|}{} &
      \multicolumn{1}{c|}{25.174/8.013/6.906} &
      1.864/1.696/1.685 &
      9/8.551/8 \\ \hline
    \multicolumn{2}{|c|}{\begin{tabular}[c]{@{}c@{}}Matching (ms)\\ (Max/Average/Min)\end{tabular}} &
      \multicolumn{1}{c|}{2.5/1.4/0.9} &
      \multicolumn{1}{c|}{0.18/0.0736/0.0701} &
      0.01/0.009/0.001 &
      1/0.28/0.10 \\ 
      \hline
    \end{tabular}
\end{center}
\end{table*}
\section{Results}
\label{section5}
In this section, we analyze the evaluation results of two groups of experiments.
We analyze and evaluate SSR-VLAD from two aspects: the characterization ability of the descriptor and the computational consumption.
\subsection{Analysis of Characterization Ability}
\label{5.1}
\begin{figure*}
	\centering
	\subfigure[]{
	    \label{fig3a} 
		\includegraphics[width=2.2in]{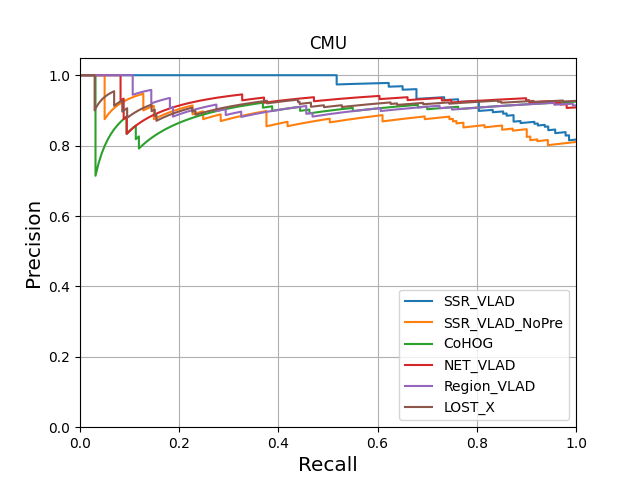}
	}
	\subfigure[]{
	    \label{fig3b} 
		\includegraphics[width=2.2in]{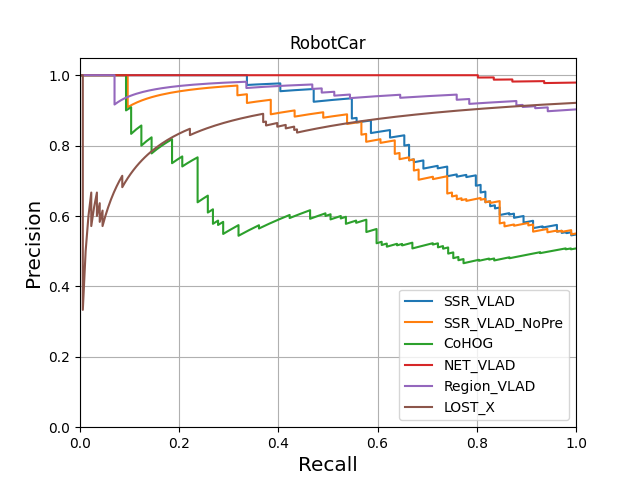}
	}
    \subfigure[]{
	    \label{fig3c} 
		\includegraphics[width=2.2in]{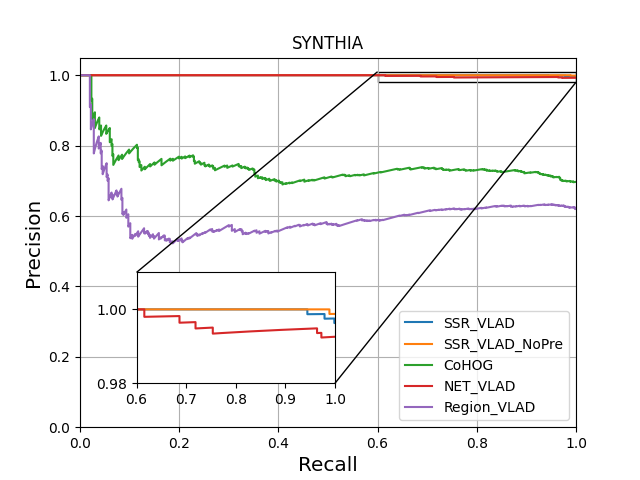}
	}
	\subfigure[]{
	    \label{fig3d} 
		\includegraphics[width=2.2in]{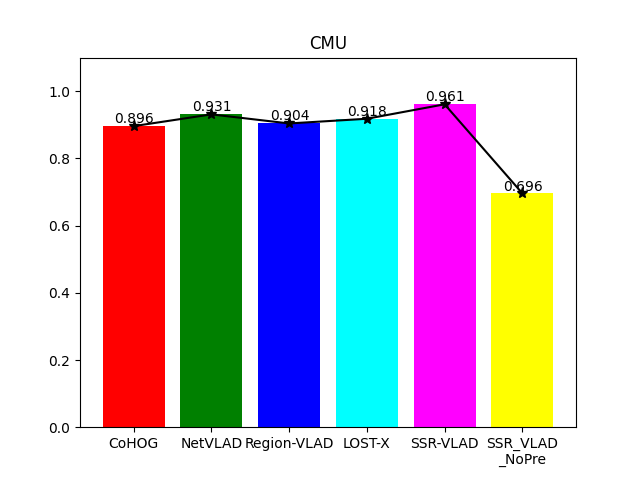}
	}
	\subfigure[]{
	    \label{fig3e} 
		\includegraphics[width=2.2in]{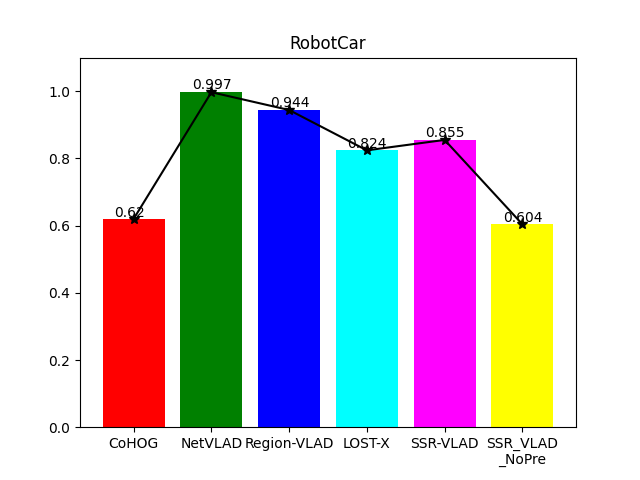}
	}
    \subfigure[]{
	    \label{fig3f} 
		\includegraphics[width=2.2in]{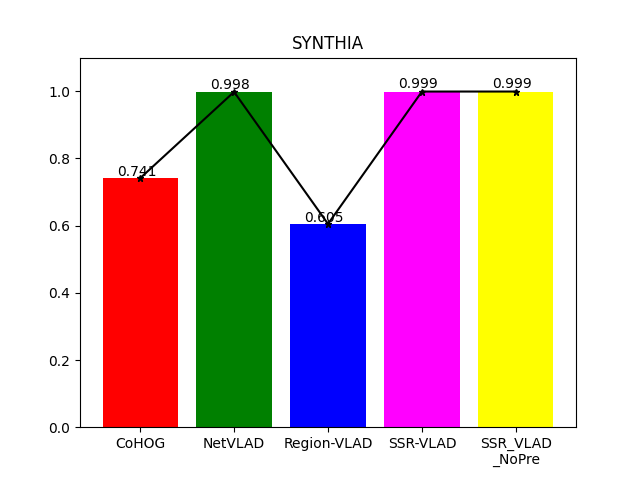}
	}
	\caption{Experiment result on the evaluated long-term urban datasets. Fig\ref{fig3a}-Fig\ref{fig3c} show the PR-curve of all evaluated methods in three datasets. Fig\ref{fig3d}-Fig\ref{fig3f} show the comparison of the AUC values of all evaluated methods in three datasets.}
	\label{fig3} 
\end{figure*}

As mentioned before, we evaluate the characterization capabilities of all compared methods on the same laptop computer.
In order to ensure the consistency of the input semantic information and the SYNTHIA dataset cannot provide the necessary feature map for LOST-X, so we cannot evaluate the performance of LOST-X on SYNTHIA in this experiment.
The evaluation metrics results of all methods on test datasets are shown in Table \ref{table1}. 
Moreover, we utilize the PR curve to intuitively represent the results of this experiment, as shown in Figure\ref{fig3}.
Experimental results show that all evaluation metrics of SSR-VLAD in all test datasets have achieved state-of-the-art. 
For AUC value, SSR-VLAD is outperformance and only 10\% worse than NetVLAD. And in SYNTHIA dataset, this metric of SSR-VLAD achieves 0.999804 (1 is the best). 
This means that SSR-VLAD has balanced precision and recall performance for VPR task.
For Recall@100\%Precision, although SSR-VLAD is lower than Net-VLAD, it is more than 80\% better than the other three methods.
Therefore, SSR-VLAD could retrieve more correct revisited candidate than CoHOG, Region-VLAD and LOST-X.
For Recall@1, SSR-VLAD achieves 99.6\%(100\% is the best) in SYNTHIA but is worse than other methods in CMU and RobotCar datasets.
This makes SSR-VLAD unable to detect as many revisited candidates as other methods under the same condition.
Moreover, we can find that for datasets whose segmentation results are not ideal, such as CMU and RobotCar, pre-processing the data before extracting the descriptor can effectively improve the robustness of SSR-VLAD. For the SYNTHIA dataset with accurate pixel-wise annotation, the effect of pre-processing is negative because this step breaks some original correct expressions.
Finally, compared with another semantic-based method, LOST-X, SSR-VLAD has a stronger ability to describe semantic features, which enables it to accurately distinguish true negative frames with dissimilar semantic segmentation. But this also results that it usually ignores some true positive frames whose segmentation results are very different from query image due to mis-segmentation. That is why SSR-VLAD performs much better than LOST-X on the Recall@100\%Precision metric but not as good as LOST-X on the Recall@1 metric. Combined with the comprehensive analysis of AUC value, SSR-VLAD has a more comprehensive and balanced performance than LOST-X on the test datasets.

Overall, the performance of SSR-VLAD in the CMU and SYNTHIA datasets is the best among all methods, and it is only better than CoHOG in the RobotCar dataset. We guess that this is because there are more halos and motion distortions in the test images of the RobotCar dataset, leading to a large number of incorrect semantic segmentation results. These erroneous segmentation results increase the difference of SSR-VLAD between the query image and the true matched reference image.
Furthermore, SSR-VLAD presents better performance than CoHOG in all three datasets. This illustrates that semantic feature is more robust than appearance-based methods toward changing urban environment.
Moreover, SSR-VLAD achieves comparable performance with Net-VLAD, which is a network specially trained for VPR task. 
Therefore, we can find that SSR-VLAD, which are extracted based on semantic segmentation, can be applied to the VPR task of urban scenes with changes in appearance, and can achieve the SoTA performance.
\subsection{Analysis of Computational Cost}

The computational cost of a VPR algorithm at run-time is also an important metric for evaluating.
The results of computational cost experiments are shown in Table\ref{table3}. 
Since the CoHOG algorithm generates descriptors while extracting features, for CoHOG we merge the time-consuming process of feature extraction and descriptor encoding. 
We could find that the results of CoHOG and Region-VLAD shown in Table \ref{table1} and Table \ref{table3} are different from the data presented in \cite{zaffar2020cohog} and \cite{Region-VLAD}.
The reason is that the computing power of the experimental platform used in this paper is far lower than theirs and when the code is running, all cores of the CUP of the experimental platform are working at full capacity.
In addition, we suppose that the semantic segmentation results are pre-provided. The feature extraction cost of VLAD does not include the cost of this part. 
Because the semantic segmentation module can be decoupled from the system, the semantic segmentation map is used as the default initial input of SSR-VLAD, and other semantic-related works also follows this suppose\cite{guo2021semantic} when evaluating the computational cost.
In the experiments of this paper, we use the same platform and run each method separately to ensure the fairness of the experimental results.

First of all, we can find that encoding the SSR-VLAD of one image only need to cost about $150 ms$.
While performing matching, $1 ms$ could match more than $100$ pairs of image.
Moreover, we can find that the time-consuming of SSR-VLAD is more stable than compared methods. 
This is mainly because that SSR-VLAD is designed as a fixed-dimensional descriptor, no matter how complicated the scene is.
Finally, we focus on the time-consuming performance of SSR-VLAD on embedded devices.
The results show that due to the longer feature extraction time cost, SSR-VLAD can be applied to an offline mapping and online positioning system or a low-speed online SLAM system.

\begin{table}[]
\label{table4}
\caption{Experimental results of the robustness analysis \\ for semantic skeleton}
\begin{center}
\begin{tabular}{ccccccc}
\hline
\multirow{2}{*}{}                                           & \multicolumn{6}{c}{Noise(Pixel)} \\
& $\pm25$   & $\pm50$   & $\pm75$   & $\pm100$  & $\pm125$  & $\pm150$    \\ \hline
\begin{tabular}[c]{@{}c@{}}Average Recall@1\\  / 20 times\end{tabular} 
& 1    & 0.989  & 0.914  & 0.853  & 0.797 & 0.663  \\ \hline
\end{tabular}
\end{center}
\end{table}
\subsection{Robustness analysis of semantic skeleton graph}
According to the experimental results shown in Table \ref{table4}, we can find that the semantic skeleton graph has the ability to resist certain segmentation noise, but the excessive noise will cause the representation ability of the skeleton graph to decrease. Combined with the conclusion analysis of Section\ref{5.1}, it can be considered that the characterization ability of SSR-VLAD is positively correlated with the effect of semantic segmentation.
\section{Conclusion}
In this paper, a lightweight semantic image descriptor, SSR-VLAD, is proposed. 
SSR-VLAD considers the shape, category and spatial distribution information of semantic elements in the image and semantic consistency between adjacent images.
Through encoding the above information in space and time sequence, SSR-VLAD can describe the feature of environment under the appearance variation caused by seasons and illumination.
Experimental results show that SSR-VLAD exhibits state-of-the-art scene characterization ability under urban scenes with drastic appearance variation. In addition, through comparison with two other outstanding lightweight models, SSR-VLAD shows very low computing power consumption on general laptop platforms. Moreover, SSR-VLAD can be used conditionally on embedded devices.
Due to commercial issues, this work cannot be directly open sourced. Colleagues who are interested in this work can contact us by e-mail to obtain the code conditionally.

\section{Acknowledgement}
This work was supported by the Research and development of L3 commercial vehicle autonomous driving system based on artificial intelligence, [grant numbers XLYC1902029], and Research and development and industrialization of key technologies for intelligent driving vehicles, [grant numbers 2019JH1/10100026].

\bibliographystyle{IEEEtran}
\bibliography{root}

\end{document}